\documentclass[letterpaper, 10 pt, journal, twoside]{IEEEtran}
\usepackage{graphicx}
\usepackage{cite}
\usepackage{url}
\usepackage{colortbl}
\usepackage{multirow}
\usepackage{amssymb}
\usepackage{mathtools}
\usepackage{bbm}
\usepackage{xcolor}
\usepackage{booktabs}

\usepackage{hyperref}
\hypersetup{
    pdftitle={PhysReal: Learning Real-World Deformable Object Physics via Hybrid Constitutive Modeling},
    pdfauthor={Yinan Deng, Jianqiao Song, Yisi Zhang, Yuhan Wang, Jiahui Wang, and Yufeng Yue},
    colorlinks=true,
    linkcolor=black,
    filecolor=magenta,      
    urlcolor=blue,
}

\definecolor{limegreen}{rgb}{0.2, 0.8, 0.2}
\definecolor{forestgreen}{rgb}{0.13, 0.55, 0.13}
\definecolor{greenhtml}{rgb}{0.0, 0.5, 0.0}
\definecolor{black}{rgb}{0.0, 0.0, 0.0}

\usepackage[font=small,labelfont=bf,tableposition=top]{caption}

\usepackage{blindtext}

\begin{document}

\title
{
PhysReal: Learning Real-World Deformable Object Physics via Hybrid Constitutive Modeling
}

\author{
Yinan~Deng$^{1}$, Jianqiao~Song$^{1}$, Yisi~Zhang$^{1}$,
Yuhan~Wang$^{1}$, Jiahui~Wang$^{1}$, and Yufeng~Yue$^{1*}$%
\thanks{Manuscript received: August 24, 2026; Revised: November 3, 2026; Accepted: November 30, 2026.}%
\thanks{This paper was recommended for publication by Editor Teresa A. Vidal-Calleja upon evaluation of the Associate Editor and Reviewers' comments.}
\thanks{This work is supported by the National Natural Science Foundation of China under Grant 625B2024, 62473050, 92370203. \textit{(Corresponding author: Yufeng Yue.)}}%
\thanks{$^{1}$All authors are with the School of Automation, Beijing Institute of Technology, Beijing 100081, China (e-mail: yueyufeng@bit.edu.cn).}%
\thanks{Digital Object Identifier (DOI): see top of this page.}
}

\markboth{IEEE Robotics and Automation Letters. Preprint Version. Accepted November, 2026}%
{Deng \MakeLowercase{\textit{et al.}}: PhysReal}

\maketitle

\begin{abstract}

\textcolor{black}{Learning physically plausible object dynamics from visual observations is important for robotic manipulation and interactive simulation.} However, modeling real-world deformable objects remains challenging because their dynamics often arise from complex, spatially heterogeneous material responses.
To address this challenge, we propose PhysReal, a video-driven framework for learning and simulating the underlying physics of real deformable objects.
PhysReal integrates a spatially varying hybrid expert-neural constitutive model with a differentiable MPM simulator and 3DGS renderer. Analytical expert models provide interpretable physical priors, while neural constitutive residuals capture material responses beyond predefined formulations. \textcolor{black}{Spatially distributed patches parameterize the constitutive field, enabling a compact representation of local material variations.}
To organize the identification of this model from sparse visual observations, we adopt a progressive curriculum that sequentially optimizes global material properties, spatially varying local parameters, and neural constitutive residuals, together with complementary motion and mask supervision.
\textcolor{black}{Extensive experiments on diverse deformable-object interactions demonstrate that PhysReal achieves superior performance in dynamic reconstruction and future-state prediction.}
Our \href{https://physreal.github.io/project_web}{project~website} is publicly available.

\end{abstract}

\begin{IEEEkeywords}
Deformable Object Manipulation, Physics Simulation, \textcolor{black}{System Identification.}
\end{IEEEkeywords}

\section{Introduction}

\IEEEPARstart{U}{nderstanding} and modeling physical interactions is fundamental to intelligent agents operating in the real world. While end-to-end video prediction models can generate visually plausible motion, recovering the underlying physical dynamics requires identifying the material responses that govern object deformation. This is particularly challenging for deformable objects, whose behavior depends on complex deformation mechanisms and spatially varying material properties \cite{Phystwin, EMPM, PGND}.

Existing approaches to deformable object modeling fall into two broad categories. Learning-based methods leverage neural networks, such as GNNs \cite{Adaptigraph, SoMA, GSDynamic} or MLPs \cite{PGND}, to directly learn deformation dynamics from data, yet often lack interpretability and explicit physical constraints, limiting their generalization to novel interaction patterns.
Physics-based methods model deformation dynamics based on mechanical principles. Existing methods typically rely on discrete formulations, such as Spring-Mass Models (SMMs) and Position-Based Dynamics (PBD), or continuum-based formulations, such as Finite Element Methods (FEM) and the Material Point Method (MPM). 
\textcolor{black}{Among these formulations, spring models capture local compliance through spatially varying stiffness \cite{Phystwin, Spring-Gaus, Neuspring}. However, the recovered stiffnesses characterize discrete connections rather than directly specifying continuum material moduli. 
In contrast, continuum formulations describe material responses through constitutive relations between stress and strain.
However, most of these approaches are constrained by predefined analytical constitutive models and assume homogeneous material properties, limiting their
applicability to complex real-world deformable objects.}
This naturally raises the question: \textbf{how can we learn the rich underlying physical dynamics of real deformable objects?}
To answer this question, we identify two key challenges.

\begin{figure}[!t]\centering
	\includegraphics[width=8.5cm]{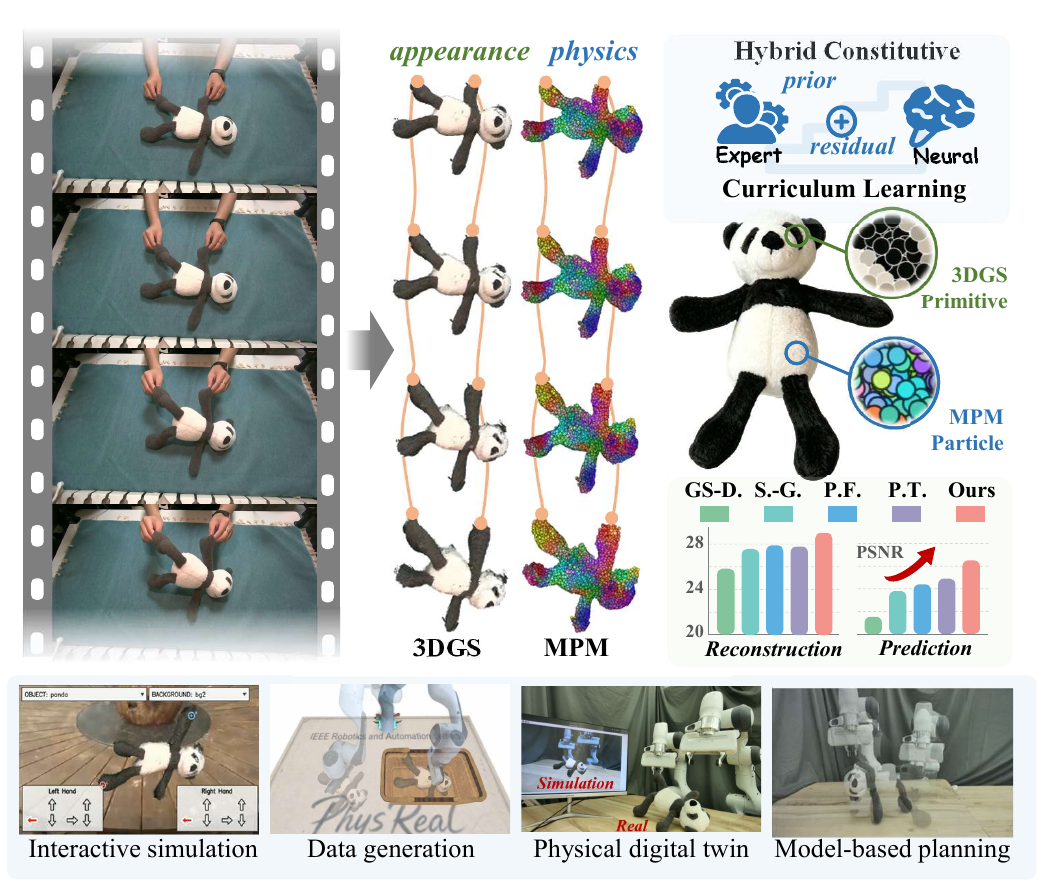}
    \caption{\textcolor{black}{Given a single-view interactive video, PhysReal couples 3DGS-based appearance with differentiable MPM physics and learns object-specific deformable dynamics through spatially varying hybrid constitutive fields. 
Downstream examples illustrate potential uses of the recovered model.}}
    \label{first}
    \vspace{-5pt}
\end{figure}

\textbf{I. Stably modeling complex spatially varying constitutive responses.}
\textcolor{black}{Real deformable objects often exhibit locally varying material properties, leading to heterogeneous deformation behaviors that are difficult to capture with a fixed constitutive formulation. Recent differentiable physics approaches estimate spatially varying material parameters or locally select analytical constitutive models within MPM \cite{Physdreamer, Physworld, Omniphysgs}. However, parameter estimation and model selection remain restricted to predefined constitutive families and cannot capture material responses beyond those families. Meanwhile, concurrent neural residual approaches improve deformation prediction by correcting simulated motion \cite{DeformMaster, PhysCoRe}, but these corrections act on particle states rather than explicitly modeling the underlying constitutive responses.}
To address this limitation, we propose a spatially varying hybrid expert-neural constitutive model, where analytical constitutive models provide interpretable physical priors and neural constitutive residuals capture unmodeled material behaviors. 
\textcolor{black}{Learnable material parameters and latent codes assigned to local patches further represent spatial variations in both components.}

\textbf{II. Efficiently learning differentiable physics from sparse visual observations.}
Learning spatially varying constitutive representations from single-view sparse visual observations is challenging because videos provide only indirect observations of physical states \cite{MASIV,OmniMap}. Moreover, jointly estimating the expert and neural components creates a highly coupled optimization problem with potentially ambiguous solutions.
\textcolor{black}{We therefore progressively optimize global and local expert parameters, then fix them while learning neural residuals and local latent codes.} Furthermore, we introduce complementary motion and mask supervision by combining tracker-based motion constraints for dynamic fitting and differentiable 3DGS-based mask supervision for accurate silhouette refinement.

In summary, as illustrated in Fig.~\ref{first}, we present PhysReal, a deformable object modeling and simulation framework that learns realistic physical behaviors from single-view interactive
videos.
The contributions are summarized as follows:

\begin{itemize}
    \item We introduce PhysReal, a novel video-driven framework that recovers deformable dynamics of real-world objects with complex and heterogeneous material behaviors.
    \item \textcolor{black}{We propose a patch-conditioned hybrid constitutive model combining analytical elastic stresses and plastic updates with neural residuals for complex responses.}
    \item \textcolor{black}{We develop a curriculum learning strategy with complementary motion and mask supervision for recovering physical properties from single-view visual observations.}
    \item  \textcolor{black}{Extensive experiments demonstrate superior performance in reconstruction and future prediction, complemented by preliminary explorations of robotic applications.}
\end{itemize}

\begin{figure*}[!t]\centering
	\includegraphics[width=0.9\textwidth]{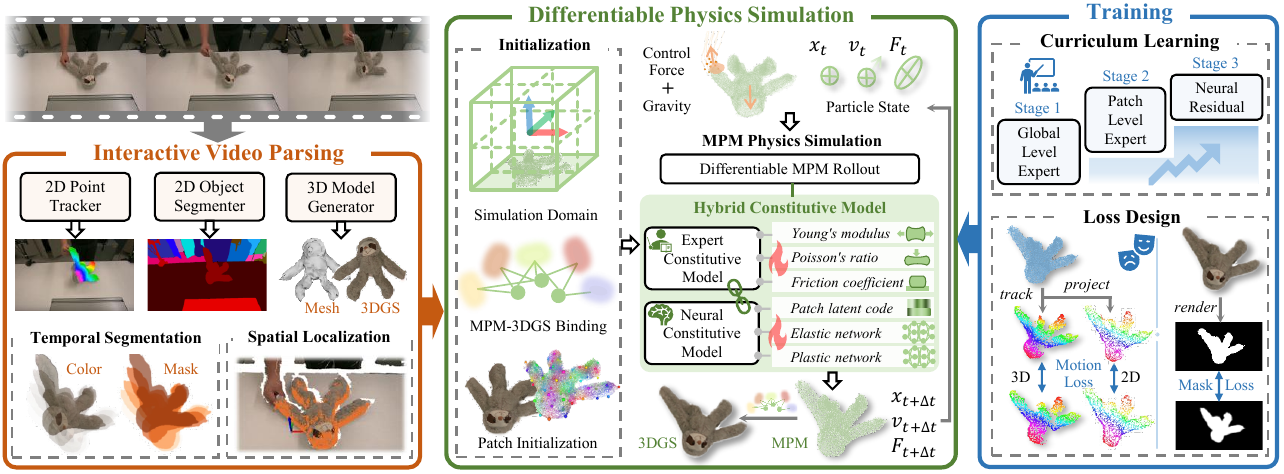}
	\caption{Overview of PhysReal. The framework consists of three key components: \textcolor{black}{(1) \textbf{Interactive Video Parsing}, which performs temporally consistent object segmentation and 3D spatial localization from interactive videos; (2) \textbf{Differentiable Physics Simulation}, which integrates MPM with a patch-conditioned hybrid expert-neural constitutive model,} and (3) \textbf{Training}, which adopts a curriculum strategy with complementary motion and mask supervision.}
	\label{framework} 
    \vspace{-5pt}
\end{figure*}

\section{Related Work} \label{RW}

\subsection{Learning-based Deformable Dynamics Modeling}

Learning-based approaches infer deformable dynamics directly from data. A common paradigm represents objects as structured particles, graphs, or meshes and learns their state transitions through neural message passing. DPI-Nets \cite{DPI-Nets} learns particle interactions for deformable objects and fluids, GNS \cite{GNS} performs dynamics propagation over particle graphs, and MeshGraphNets \cite{MeshGraphNets} extends graph-based simulation to mesh representations. 

Within this paradigm, one line of work focuses on adapting learned dynamics to object-specific materials and visual observations \cite{PaMoSplat}. Adaptigraph \cite{Adaptigraph} conditions graph dynamics on material information, GSDynamic \cite{GSDynamic} couples dynamic 3D Gaussian tracking with graph-based prediction, and PGND \cite{PGND} combines particle and grid representations to infer deformation dynamics from RGB-D videos. \textcolor{black}{SoMA \cite{SoMA} develops a real-to-sim neural simulator for embodied soft-body manipulation.}

\textcolor{black}{Despite their flexibility, these methods encode material responses implicitly in learned state transitions rather than explicit constitutive relations, which limits the physical interpretability of the learned dynamics and can hinder generalization to interactions beyond the training distribution.}

\subsection{Physics-based Deformable Simulation}

Physics-based methods instead employ explicit mechanical models for deformable simulation, providing greater physical interpretability and consistency. One class of methods relies on SMMs, where deformation is represented by particle-spring systems. \textcolor{black}{Spring-Gaus \cite{Spring-Gaus} and NeuSpring \cite{Neuspring} combine spring models with Gaussian or neural representations, while PhysTwin \cite{Phystwin} optimizes dense spring stiffness from interaction videos to represent spatially varying compliance. However, the recovered stiffnesses characterize discrete spring connections, limiting their direct interpretation as intrinsic material moduli.}

\textcolor{black}{Another class adopts continuum simulation, especially MPM, for deformable dynamics modeling. PhysGaussian \cite{Physgaussian} couples MPM with 3DGS for dynamics generation, while GIC \cite{Gic} targets material identification. EMPM \cite{EMPM} extends MPM to embodied interaction, and PhysWorld \cite{Physworld} optimizes global and local material parameters. OmniPhysGS \cite{Omniphysgs} captures spatial variation through local selection among analytical models. Such parameter estimation and model selection remain restricted to predefined constitutive families. NCLaw \cite{NClaw} instead learns elastic and plastic laws with neural networks. However, learning the entire constitutive response without an analytical backbone can be difficult under sparse visual supervision.
Concurrent methods DeformMaster \cite{DeformMaster} and PhysCoRe \cite{PhysCoRe} introduce neural residuals to compensate for motion discrepancies, but these corrections act on simulated states rather than within constitutive laws.}

In contrast, PhysReal learns deformable objects as spatially varying constitutive fields by combining expert physical priors with neural constitutive residuals. Moreover, PhysReal introduces progressive curriculum learning and complementary motion and mask supervision to stabilize the optimization of complex physical models from interactive videos.

\section{PhysReal}  \label{method}

\subsection{Framework Overview}
\label{sec:framework}

\textbf{Problem Definition.}
Given an interactive RGB-D observation sequence $\mathcal{O}=\{\mathbf{C}_t, \mathbf{D}_t\}$, PhysReal aims to learn a physically grounded deformable object model that can reproduce and predict object dynamics under an external action sequence $\mathcal{A}=\{\mathbf{a}_t\}$.

We represent the deformable object as a set of material particles following MPM. The physical state at time $t$ is defined as $\mathbf{s}_t$. The objective is to recover a differentiable physical transition function $f_{\Theta}$:
\begin{equation}
    \mathbf{s}_{t+1}
    =
    f_{\Theta}
    (\mathbf{s}_{t},\mathbf{a}_{t}),
    \label{eq:state_transition}
\end{equation}
where $\Theta$ represents the learnable physical parameters, including material properties and neural representations.

As illustrated in Fig. \ref{framework}, PhysReal consists of three main components. \textcolor{black}{First, the \textit{Interactive Video Parsing} module extracts temporally consistent object observations from interaction videos. Second, the \textit{Differentiable Physics Simulation} module constructs an MPM-based physical simulator and introduces a spatially varying hybrid expert-neural constitutive model.} Finally, the \textit{Training} module progressively optimizes global material properties, local constitutive variations, and neural residual responses through curriculum learning with complementary motion and mask supervision.

\subsection{Interactive Video Parsing}
\label{sec:video_parsing}

\textbf{Temporally Consistent Object Segmentation and Motion Tracking.}
Given a single-view RGB-D video sequence, we first employ an off-the-shelf instance segmentation model \cite{entityv2} to obtain candidate masks $\textcolor{black}{\mathcal{Q}_{t}}$ in each frame. For the reference frame $\mathbf{C}_0$, the target object is selected according to semantic correspondence between the object description and segmentation candidates \cite{video2robo}, producing the initial object mask $\mathbf{M}_0$. \textcolor{black}{We sample points from the initial object region and employ a pretrained pixel tracker \cite{Cotracker3} to establish temporal correspondence across frames.}
The obtained motion trajectories are represented as:
\begin{equation}
    \mathcal{T}
    =
    \{
    \mathbf{u}^{j}_{t},\omega_t^j
    \},
    \label{eq:motion_track}
\end{equation}
where $\mathbf{u}^{j}_{t}\in\mathbb{R}^{2}$ denotes the image coordinate of pixel $j$ at frame $t$, and $\omega_t^j$ indicates the visibility of the tracked point. 

To obtain temporally consistent object masks \cite{Openobj}, we associate tracked points with segmentation candidates in each frame. Specifically, the candidate mask receiving the highest number of valid tracked points is selected:
\begin{equation}
    \mathbf{M}_{t}
    =
    \arg\max_{\mathbf{m}\in\textcolor{black}{\mathcal{Q}_{t}}}
    \sum_{j}
    \omega_t^j
    \mathbb{I}
    (\mathbf{u}^{j}_{t}\in\mathbf{m}),
    \label{eq:mask_tracking}
\end{equation}
where $\textcolor{black}{\mathcal{Q}_{t}}$ denotes the candidate masks at time $t$. The resulting mask sequence $\mathcal{M}=\{\mathbf{M}_t\}$ provides geometric supervision for later optimization.

\textbf{3D Object Generation and Spatial Localization.}
Besides temporal observations, physical simulation requires an object-centric geometric representation aligned with the real scene.
We leverage a generative model \cite{TRELLIS} to reconstruct the object geometry from the reference observation $\mathbf{C}_0[\mathbf{M}_0]$, yielding an initial 3DGS representation $\mathcal{G}_0$ and mesh model $\mathcal{S}_0$.
Since the generated model is defined in an arbitrary canonical coordinate system, we further optimize its spatial transformation with respect to the observed RGB-D $\{\mathbf{C}_0, \mathbf{D}_0\}$.

The transformation between the generated object and camera coordinate system is parameterized as:
\begin{equation}
    \boldsymbol{\xi}
    =
    \textcolor{black}{(\mathbf{R},\mathbf{b},\alpha)},
\end{equation}
\textcolor{black}{where $\mathbf{R}$, $\mathbf{b}$, and $\alpha$ denote rotation, translation, and scale, respectively.} We formulate the localization process as an optimization problem:
\begin{equation}
    \boldsymbol{\xi}^{*}
    =
    \arg\min_{\boldsymbol{\xi}}
    (\textcolor{black}{\mathcal{L}_{\mathrm{rgb}}^{\mathrm{loc}}}
    +
    \textcolor{black}{\mathcal{L}_{\mathrm{depth}}^{\mathrm{loc}}}
    +
    \textcolor{black}{\mathcal{L}_{\mathrm{mask}}^{\mathrm{loc}}}),
    \label{eq:localization_loss}
\end{equation}
\textcolor{black}{where $\mathcal{L}_{\mathrm{rgb}}^{\mathrm{loc}}$, $\mathcal{L}_{\mathrm{depth}}^{\mathrm{loc}}$, and $\mathcal{L}_{\mathrm{mask}}^{\mathrm{loc}}$ are the reference-frame RGB, depth, and mask losses for initial localization, respectively. They compare renderings of $\mathcal{G}_0$ with observations $\mathbf{C}_0$, $\mathbf{D}_0$, and $\mathbf{M}_0$.}
To handle the large non-convex search space, we first obtain a coarse estimation through Differential Evolution and then refine the transformation using differentiable rendering. 

\subsection{Differentiable Physics Simulation}
\label{sec:physics}


\textbf{Particle-based Physical Initialization.}
After aligning the object 3D models $\mathcal{S}_0$ and $\mathcal{G}_0$ with the observed scene, we first establish the physical simulation coordinate system. Specifically, the supporting surface is estimated from the reconstructed scene point cloud, and its normal direction is used to define the vertical axis of the world coordinate system. A bounded simulation domain is then constructed around the target object for MPM simulation.

To represent the object as a volumetric continuum, we sample material particles inside the generated mesh $\mathcal{S}_0$. The initial particle state is defined as:
\begin{equation}
    \mathbf{s}_0
    =
    \{
    \mathbf{x}^i_0,
    \mathbf{v}^i_0,
    \mathbf{F}^i_0
    \},
    \label{eq:particle_initialization}
\end{equation}
\textcolor{black}{where $\mathbf{x}^i_0$, $\mathbf{v}^i_0$, and
$\mathbf{F}^i_0$ denote the initial position, velocity, and deformation gradient of particle $i$.} 

\textbf{Particle-Gaussian Binding.}
To establish the connection between physical simulation and visual observations, each Gaussian primitive is associated with multiple nearby MPM particles in the initial configuration, and its deformation is determined by their weighted motion through an LBS-like deformation scheme. Given the initial Gaussian representation $\mathcal{G}_0$, the initial particle state $\mathbf{s}_0$, and the evolved particle state $\mathbf{s}_t$, the deformed Gaussian representation is obtained as
\begin{equation}
    \mathcal{G}_t
    =
    \mathcal{B}
    (
    \mathcal{G}_0,
    \mathbf{s}_0,
    \mathbf{s}_t
    ),
    \label{eq:gs_binding}
\end{equation}
where $\mathcal{B}$ denotes the particle-to-Gaussian deformation mapping defined by the initial binding and the particles' subsequent motion. The deformed Gaussian representation is then rendered into the 2D image space:
\begin{equation}
    \textcolor{black}{\hat{\mathbf{C}}_t},
    \hat{\mathbf{D}}_t,
    \hat{\mathbf{M}}_t
    =
    \mathcal{R}
    (
    \mathcal{G}_t
    ),
    \label{eq:physics_rendering}
\end{equation}
where $\textcolor{black}{\hat{\mathbf{C}}_t}$, $\hat{\mathbf{D}}_t$, and $\hat{\mathbf{M}}_t$ denote the rendered RGB image, depth image, and alpha mask, respectively. 

\textbf{Spatially Varying Constitutive Field.}
\textcolor{black}{We represent the material field compactly using patches $\mathcal{P}=\{\mathbf{p}^k\}$ selected by farthest point sampling of the initial particles. Each patch stores}
\begin{equation}
    \boldsymbol{\theta}^k
    =
    \{
    E^k,\nu^k,
    \mathbf{z}^{k,e},
    \mathbf{z}^{k,p}
    \},
    \label{eq:patch_representation}
\end{equation}
where $E^k$ and $\nu^k$ denote the local Young's modulus and Poisson's ratio, while $\mathbf{z}^{k,e}$ and $\mathbf{z}^{k,p}$ represent elastic and plastic latent codes for neural constitutive modeling.

\textcolor{black}{Patch representations are interpolated to each particle as}
\begin{equation}
    \boldsymbol{\theta}^i
    =
    \mathcal{I}
    (
    \{\boldsymbol{\theta}^k\},
    \{\mathbf{p}^k\},
    \mathbf{x}^{i}_{0}
    )
    =
    \{E^i,\nu^i,\mathbf{z}^{i,e},\mathbf{z}^{i,p}\},
    \label{eq:patch_to_particle}
\end{equation}
\textcolor{black}{where $\mathcal{I}$ averages the parameters and latent codes of the three nearest patches using normalized inverse-distance weights. Neighbors and weights are fixed at initialization, keeping the material field attached to the deforming object.}

\textbf{Action-driven Physical Interaction.}
\textcolor{black}{The action sequence $\mathcal{A}=\{\mathbf{a}_t\}$ comprises 3D hand or robot interaction-point trajectories, interpolated between video frames. Each point is initially bound to nearby particles with Gaussian distance weights $w^i$. For particle $i$ with initial binding distance $\ell^i_0$, the compliant actuation force is}
\begin{equation}
\mathbf{f}^{i,\mathrm{act}}_{t}
=
-k_f w^i
\left(
\|\mathbf{x}^{i}_{t}-\mathbf{a}_t\|/\textcolor{black}{\ell^i_0}-1
\right)
\mathbf{d}^{i}_{t},
\label{eq:interaction_force}
\end{equation}
\textcolor{black}{where $k_f$ is the actuation stiffness and $\mathbf{d}^{i}_{t}$ points from $\mathbf{a}_t$ to $\mathbf{x}^{i}_{t}$. Particle motion follows the resulting forces and constitutive dynamics rather than prescribed positions.}

\textbf{Hybrid Expert-Neural Constitutive Model.}
For each particle, the expert elastic response is computed from its deformation state and local material parameters:
\begin{equation}
    \textcolor{black}{\boldsymbol{\tau}^{i,\mathrm{exp}}_t}
    =
    \mathcal{C}_{e}
    (
    \mathbf{F}^i_t,
    E^i,\nu^i
    ),
\end{equation}
\textcolor{black}{where $\mathbf{F}^i_t$ is the elastic deformation gradient at the start of the substep, $\boldsymbol{\tau}^{i,\mathrm{exp}}_t$ is the expert Kirchhoff stress, and $\mathcal{C}_e$ is the analytical elastic law.}
\textcolor{black}{An elastic residual network, shared across particles and conditioned on their local elastic codes, captures responses beyond the analytical model:}
\begin{equation}
    \textcolor{black}{\Delta\boldsymbol{\tau}^{i,\mathrm{neu}}_t}
    =
    \mathcal{N}_{e}
    (
    \textcolor{black}{\mathbf{F}^i_t},
    \mathbf{z}^{i,e}
    ),
\end{equation}
\textcolor{black}{Here $\mathbf{z}^{i,e}$ is interpolated from the patches.}
The final elastic stress used in MPM is obtained by:
\begin{equation}
    \boldsymbol{\tau}^i_t
    =
    \textcolor{black}{\boldsymbol{\tau}^{i,\mathrm{exp}}_t}
    +
    \textcolor{black}{\Delta\boldsymbol{\tau}^{i,\mathrm{neu}}_t} .
\end{equation}

\textcolor{black}{Using $\boldsymbol{\tau}^i_t$ together with external forces, one MPM substep produces the trial elastic deformation gradient $\widetilde{\mathbf{F}}^i_{t+\Delta t}$. The expert plastic model then updates this trial state as}
\begin{equation}
\mathbf{F}^{i,\mathrm{exp}}_{t+\Delta t}
=
\mathcal{C}_p(\widetilde{\mathbf{F}}^i_{t+\Delta t}),
\label{eq:plastic_expert}
\end{equation}
\textcolor{black}{where $\mathcal{C}_p$ returns the elastic gradient after the expert plastic update. For identity plasticity, the trial gradient is unchanged. The neural branch uses the same trial gradient:}
\begin{equation}
    \Delta\mathbf{F}^{i,\mathrm{neu}}_{t+\Delta t}
    =
    \mathcal{N}_{p}
    (
    \textcolor{black}{\widetilde{\mathbf{F}}^i_{t+\Delta t}},
    \mathbf{z}^{i,p}
    ),
    \label{eq:neural_plastic_residual}
\end{equation}
\textcolor{black}{The plastic branch uses the local code $\mathbf{z}^{i,p}$ to predict a correction to the elastic deformation gradient. The corrected gradient is}
\begin{equation}
    \mathbf{F}^{i}_{t+\Delta t}
    =
    \mathbf{F}^{i,\mathrm{exp}}_{t+\Delta t}
    +
    \Delta\mathbf{F}^{i,\mathrm{neu}}_{t+\Delta t}
    .
\label{eq:hybrid_plastic_update}
\end{equation}

In addition to the constitutive parameters, we optimize a global friction coefficient $\mu_f$ to model object-ground interactions during physical simulation.

\textcolor{black}{\textbf{Physical Structure and Numerical Stability.} Within each residual network, the elastic deformation gradient is encoded using its principal stretches, right Cauchy-Green tensor, and determinant, centered at their undeformed values. These features are invariant under superposed rigid rotations. The polar rotation of the input gradient restores spatial orientation to symmetrized network outputs, and the elastic branch constructs a symmetric Kirchhoff stress correction. Centered features, networks without biases, and zero local codes yield zero reference corrections at initialization. Material bounds and clamping the determinant used in expert stress evaluation help control numerical extremes. Elastic residual entries are bounded relative to a regularized expert stress scale using $\beta_e$, with an additional cap on total stress. Plastic corrections are scaled, have their trace removed, and are bounded in norm relative to the trial elastic gradient using $\beta_p$. A sigmoid gate based on the correction norm attenuates small plastic updates to limit their accumulation. Qualitative tests with the simulator running continuously for 24 hours of wall-clock time showed no visible deformation drift. The objects fell freely under gravity after the actuated regions were released.}

\textbf{Differentiable Physical Rollout.}
\textcolor{black}{MPM integrates constitutive, actuation, gravity, and contact effects through particle-to-grid transfer, grid evolution, and grid-to-particle transfer. The resulting particle-state evolution is summarized as}
\begin{equation}
\mathbf{v}^{i}_{t+\Delta t}
=
\mathbf{v}^{i}_{t}
+
\Delta t\frac{\mathbf{f}^{i}_{t}}{m_i},
\quad
\mathbf{x}^{i}_{t+\Delta t}
=
\mathbf{x}^{i}_{t}
+
\Delta t\,\mathbf{v}^{i}_{t+\Delta t}.
\label{eq:state_integration}
\end{equation}
\textcolor{black}{Here $m_i$ is particle mass and $\mathbf{f}^{i}_{t}$ is the effective net force; multiple differentiable MPM substeps $\Delta t$ separate adjacent video observations.}

\subsection{Training}
\label{sec:training}

\textcolor{black}{The coupling between material parameters and neural residuals makes their joint identification from visual observations challenging.}

\textbf{Progressive Constitutive Optimization.}
We organize this estimation process using a progressive curriculum learning strategy.
The optimization process is divided into three stages:

In Stage 1, we optimize a homogeneous expert constitutive model to capture the dominant deformation characteristics of the object. The learnable parameters are defined as $\Theta_1=\{E,\nu,\mu_f\}$, where $E$, $\nu$, and $\mu_f$ denote the global Young's modulus, Poisson's ratio, and friction coefficient, respectively.

In Stage 2, the global parameters obtained are propagated to all patches, and the spatial variation of material properties is optimized as $\Theta_2=\{\{E^k\},\{\nu^k\}\}$. By allowing each local region to adapt its own constitutive parameters, this stage captures heterogeneous material distributions.

\textcolor{black}{In Stage 3, the expert constitutive parameters are fixed, and the residual networks and patch codes are optimized as $\Theta_3=\{\boldsymbol{\vartheta}_e,\boldsymbol{\vartheta}_p,\{\mathbf{z}^{k,e}\},\{\mathbf{z}^{k,p}\}\}$, where $\boldsymbol{\vartheta}_e$ and $\boldsymbol{\vartheta}_p$ denote the weights of $\mathcal{N}_e$ and $\mathcal{N}_p$, respectively.}

\begin{figure*}[!t]\centering
	\includegraphics[width=0.95\textwidth]{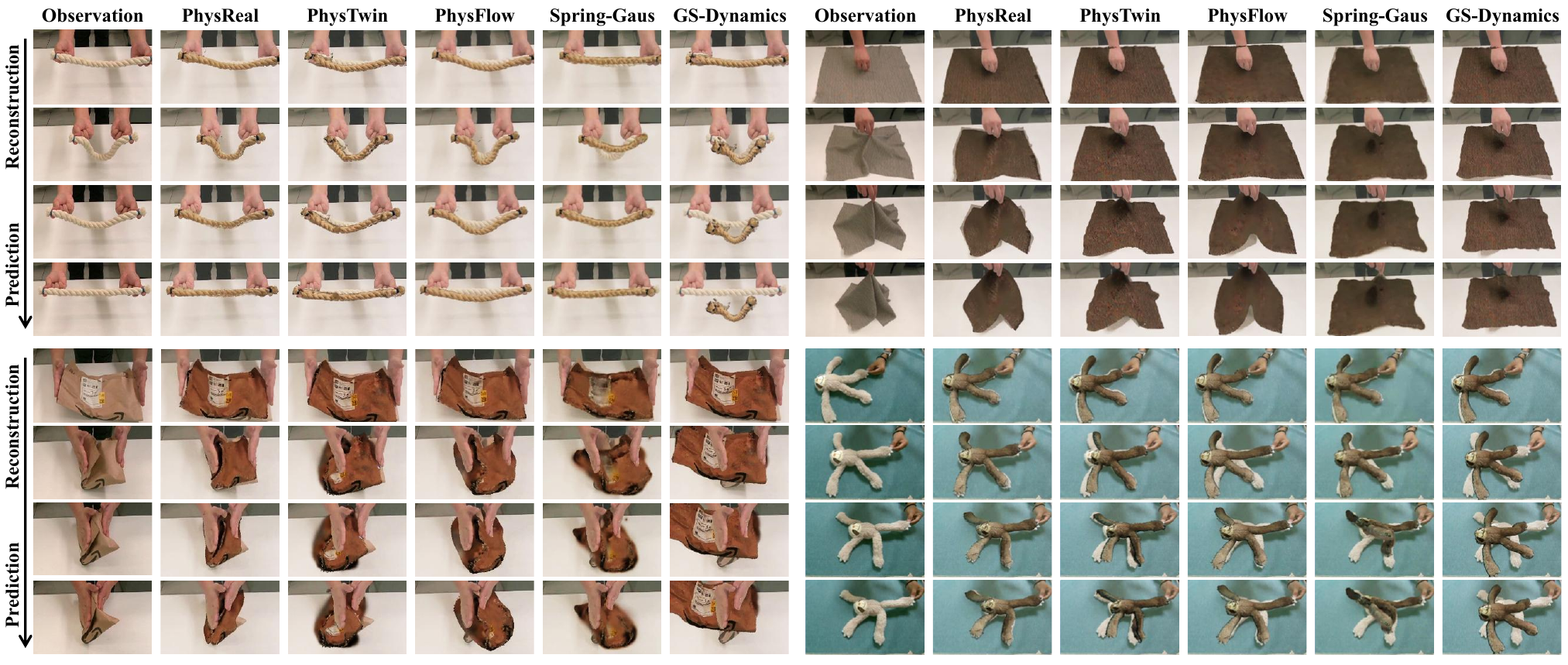}
    \caption{\textcolor{black}{Qualitative comparison on representative sequences from the PhysTwin and our self-collected datasets. For each sequence, the observed interval is used for physical model identification and reconstruction (\emph{Training}), while the remaining frames are held out for future prediction (\emph{Testing}). The same contrast adjustment is applied to all methods for clearer visualization.}}
    \label{fig:main_results}
    \vspace{-5pt}
\end{figure*}

\textbf{Complementary Motion and Mask Supervision.}
\textcolor{black}{We lift tracked pixels from Eq.~\eqref{eq:motion_track} to world coordinates using observed depth $\mathbf{D}_t$ and bind them to nearest MPM particles at initialization. Let $T$ denote the number of supervised frames. Simulated trajectories are supervised through visible 3D and 2D correspondences:}
\begin{equation}
    {\color{black}
    \mathcal{L}_{\mathrm{motion}}
    =
    \lambda_{\mathrm{3D}}\mathcal{L}_{\mathrm{3D}}
    +
    \lambda_{\mathrm{2D}}\mathcal{L}_{\mathrm{2D}}},
    \label{eq:motion_loss}
\end{equation}
\begin{equation}
{\color{black}
\mathcal{L}_{\mathrm{3D}}
=
\frac{1}{Z}
\sum_{t,j}
\omega_t^j
\rho\left(
\left\|\hat{\mathbf{x}}^j_t-\mathbf{y}^j_t\right\|_2
\right)},
\label{eq:motion_3d}
\end{equation}
\begin{equation}
{\color{black}
\mathcal{L}_{\mathrm{2D}}
=
\frac{1}{Z}
\sum_{t,j}
\omega_t^j
\rho\left(
\left\|\Pi(\hat{\mathbf{x}}^j_t)-\mathbf{u}^j_t\right\|_2
\right)},
\label{eq:motion_2d}
\end{equation}
\textcolor{black}{where $\hat{\mathbf{x}}^j_t$ is the simulated position of the particle bound to track $j$, while $\mathbf{y}^j_t$ and $\mathbf{u}^j_t$ denote the observed 3D and 2D track positions, respectively. The function $\Pi(\cdot)$ is the camera projection, $\rho(\cdot)$ is a robust penalty, and $Z=\sum_{t,j}\omega_t^j$ normalizes the visible correspondences over the supervised frames.}
\textcolor{black}{The 3D constraint $\mathcal{L}_{\mathrm{3D}}$ supervises particle trajectories in world coordinates, while $\mathcal{L}_{\mathrm{2D}}$ constrains their image-plane projections using the same tracked correspondences.}

\textcolor{black}{For dense silhouette supervision, simulated particles drive the Gaussian representation to render alpha masks $\hat{\mathbf{M}}_t$, which are compared with the temporally consistent masks from Eq.~\eqref{eq:mask_tracking}:}
\begin{equation}
    {\color{black}
    \mathcal{L}_{\mathrm{mask}}
    =
    \frac{1}{T}
    \sum_t
    \operatorname{BCE}
    (
    \hat{\mathbf{M}}_t,
    \mathbf{M}_t
    )}.
    \label{eq:mask_loss}
\end{equation}

\textcolor{black}{The physical-identification objective is}
\begin{equation}
    {\color{black}
    \mathcal{L}_{\mathrm{phys}}
    =
    \lambda_{\mathrm{motion}}\mathcal{L}_{\mathrm{motion}}
    + \lambda_{\mathrm{mask}}\mathcal{L}_{\mathrm{mask}}},
    \label{eq:training_objective}
\end{equation}
\textcolor{black}{where $\lambda_{\mathrm{motion}}$ and $\lambda_{\mathrm{mask}}$ balance sparse motion and dense silhouette supervision for constitutive identification.}
By jointly leveraging sparse motion correspondence and dense silhouette supervision, PhysReal optimizes the hidden constitutive parameters and learns physically grounded deformation dynamics from interactive videos.

\section{Experimental Results}
\label{sec:experiments}

\subsection{Experimental Setup}
\label{sec:experimental_setup}

\textcolor{black}{\textbf{Implementation Details.}
The default experts are Neo-Hookean elasticity and identity plasticity. Adapting the architecture of NCLaw~\cite{NClaw}, the elastic and plastic residual networks are separate bias-free MLPs, each with two hidden layers of 64 neurons and GELU activations. Their respective local codes are added to the first hidden activation. Each network outputs a $3\times3$ matrix used to construct the elastic stress or plastic deformation-gradient correction. We use differential evolution for global initialization and Adam with fixed stage budgets and gradient checkpointing thereafter. On an NVIDIA L40 GPU, average identification time is approximately 1\,h per object, with average peak GPU memory usage around 15\,GB.}

\textcolor{black}{\textbf{Parameter Settings.} We use 64 patches, with 64-dimensional elastic and plastic latent codes at each patch. Actuation stiffness is $k_f=1000$, and the density parameter is fixed at 1000; equal particle masses follow from density and per-particle simulation volume. Residual bounds use $\beta_e=0.1$ and $\beta_p=0.002$. The plastic output scaling factor, sigmoid-gate threshold, and transition width are each 0.001. Loss weights are $\lambda_{\mathrm{3D}}=\lambda_{\mathrm{motion}}=1$, $\lambda_{\mathrm{2D}}=0.01$, and $\lambda_{\mathrm{mask}}=2$. Global differential evolution stops at convergence or 10 generations. The local-parameter and residual stages are each optimized for 50 epochs.}

\textbf{Datasets.}
We evaluate PhysReal on both the public PhysTwin dataset \cite{Phystwin} and our self-collected interaction dataset. The PhysTwin dataset contains 22 interaction sequences covering deformable linear, planar, and volumetric objects. 
\textcolor{black}{Our self-collected dataset contains 10 interaction sequences involving four objects: a rope, a cloth, a panda plush toy, and a sloth plush toy, covering linear, planar, and volumetric geometries.}
\textcolor{black}{The first four fifths of each sequence are used for identification and reconstruction, and the remaining fifth for prediction. Metrics exclude the initialization frame. Prediction continues from the simulated state with fixed material parameters and subsequent interaction trajectories as action inputs, without observation-based state correction.}

\textbf{Baselines.}
We compare PhysReal with four representative methods: GS-Dynamics (CoRL 2024) \cite{GSDynamic}, Spring-Gaus (ECCV 2024) \cite{Spring-Gaus}, PhysFlow (CVPR 2025) \cite{physflow} and PhysTwin (ICCV 2025)  \cite{Phystwin}.
\textcolor{black}{All methods share the same initial geometry, 3DGS representation, input sequences, and temporal split. Methods originally using observations from multiple cameras are adapted by restricting supervision to the primary camera, including point tracks extracted from that view, and excluding observations from all other cameras.}

\textbf{Evaluation Metrics.}
\textcolor{black}{We report Chamfer Distance (CD $\downarrow$, m), Track Error (Track $\downarrow$, m), Intersection over Union (IoU $\uparrow$, \%), and Peak Signal to Noise Ratio (PSNR $\uparrow$, dB). CD averages bidirectional nearest neighbor distances between rendered and observed depth point clouds after removing hand occlusions and eroding masks. Track Error averages 3D Euclidean errors against CoTracker tracks lifted using observed depth, excluding invisible or invalid samples and correcting initial binding offsets. IoU thresholds rendered alpha at 0.5, while PSNR uses object images on a black background.}

\subsection{Comparison with Baselines}
\label{sec:main_results}

\textbf{Result Analysis.}
\textcolor{black}{Tabs.~\ref{table:baseline_compare}-\ref{table:baseline_compare_ours} demonstrate PhysReal's advantages in reconstruction and future prediction. On the PhysTwin dataset, it achieves the best CD, IoU, and PSNR in both settings and ties for the lowest prediction Track Error. On our self-collected dataset of complex objects, it ranks first across all metrics. These quantitative gains are consistent with the qualitative comparisons in Fig.~\ref{fig:main_results}: although most baselines generate plausible motion, PhysReal most closely reproduces the observed local deformation and overall object shape. Its hybrid constitutive field captures local compliance through spatially varying material parameters and supplements the analytical backbone with neural corrections. Together, these components support a more faithful representation of the object responses observed during interaction.}

\begin{table}[]
\centering
\caption{Quantitative comparison on the PhysTwin dataset.}
\vspace{-5pt}
\label{table:baseline_compare}
\renewcommand\arraystretch{1.0}
\setlength{\tabcolsep}{0.7mm}{
\begin{tabular}{lcccccccc}
\toprule
 & \multicolumn{4}{c}{\cellcolor[HTML]{E2EFDA}\textbf{Reconstruction}} & \multicolumn{4}{c}{\cellcolor[HTML]{D9E1F2}\textbf{Prediction}} \\
 & \cellcolor[HTML]{FFFFFF}\textbf{CD} & \cellcolor[HTML]{FFFFFF}\textbf{Track} & \cellcolor[HTML]{FFFFFF}\textbf{IoU} & \cellcolor[HTML]{FFFFFF}\textbf{PSNR} & \cellcolor[HTML]{FFFFFF}\textbf{CD} & \cellcolor[HTML]{FFFFFF}\textbf{Track} & \cellcolor[HTML]{FFFFFF}\textbf{IoU} & \cellcolor[HTML]{FFFFFF}\textbf{PSNR} \\ \midrule
\textbf{GS-Dynamics} \cite{GSDynamic} & 0.018 & 0.027 & 67.34 & 25.92 & 0.054 & 0.081 & 37.77 & 21.47 \\
\textbf{Spring-Gaus} \cite{Spring-Gaus} & 0.016 & 0.027 & 74.35 & 27.59 & 0.034 & 0.062 & 51.32 & 23.79 \\
\textbf{PhysFlow} \cite{physflow}& \underline{0.010} & 0.017 & 78.80 & \underline{27.88} & \underline{0.017} & 0.034 & 62.20 & 24.42 \\
\textbf{PhysTwin} \cite{Phystwin} & 0.011 & \textbf{0.013} & \underline{80.90} & 27.83 & \underline{0.017} & \textbf{0.027} & \underline{65.79} & \underline{24.87} \\
\textbf{PhysReal} & \textbf{0.008} & \underline{0.014} & \textbf{86.84} & \textbf{28.98} & \textbf{0.015} & \textbf{0.027} & \textbf{75.12} & \textbf{26.53} \\ \bottomrule
\end{tabular}
}
\\
\end{table}

\begin{table}[]
\centering
\caption{\textcolor{black}{Quantitative comparison on our self-collected dataset of 10 interaction sequences.}}
\vspace{-5pt}
\label{table:baseline_compare_ours}
\renewcommand\arraystretch{1.0}
\setlength{\tabcolsep}{0.7mm}{
\color{black}
\begin{tabular}{lcccccccc}
\toprule
 & \multicolumn{4}{c}{\cellcolor[HTML]{E2EFDA}\textbf{Reconstruction}} & \multicolumn{4}{c}{\cellcolor[HTML]{D9E1F2}\textbf{Prediction}} \\
 & \cellcolor[HTML]{FFFFFF}\textbf{CD} & \cellcolor[HTML]{FFFFFF}\textbf{Track} & \cellcolor[HTML]{FFFFFF}\textbf{IoU} & \cellcolor[HTML]{FFFFFF}\textbf{PSNR} & \cellcolor[HTML]{FFFFFF}\textbf{CD} & \cellcolor[HTML]{FFFFFF}\textbf{Track} & \cellcolor[HTML]{FFFFFF}\textbf{IoU} & \cellcolor[HTML]{FFFFFF}\textbf{PSNR} \\ \midrule
\textbf{GS-Dynamics} \cite{GSDynamic} & 0.019 & 0.026 & 46.16 & 21.49 & 0.045 & 0.066 & 29.88 & 18.99 \\
\textbf{Spring-Gaus} \cite{Spring-Gaus} & 0.016 & 0.022 & \underline{77.03} & \underline{25.63} & 0.027 & 0.045 & 57.85 & 21.89 \\
\textbf{PhysFlow} \cite{physflow} & \underline{0.010} & \underline{0.016} & 74.85 & 24.56 & \underline{0.017} & \underline{0.029} & \underline{65.33} & \underline{22.00} \\
\textbf{PhysTwin} \cite{Phystwin} & 0.018 & 0.022 & 51.69 & 21.76 & 0.023 & 0.034 & 47.15 & 20.68 \\
\textbf{PhysReal} & \textbf{0.008} & \textbf{0.010} & \textbf{83.55} & \textbf{26.20} & \textbf{0.013} & \textbf{0.020} & \textbf{75.33} & \textbf{23.27}
\\ \bottomrule
\end{tabular}
\vspace{-5pt}
}
\\
\end{table}

\begin{table}[]
\centering
\caption{\textcolor{black}{Prediction performance on the PhysTwin dataset grouped by object geometry.}}
\vspace{-5pt}
\label{table:type}
\renewcommand\arraystretch{1.0}
\setlength{\tabcolsep}{0.4mm}{
\begin{tabular}{lccccccccc}
\toprule
\multicolumn{1}{l}{} & \multicolumn{3}{c}{\cellcolor[HTML]{E2EFDA}\textbf{Linear}} & \multicolumn{3}{c}{\cellcolor[HTML]{D9E1F2}\textbf{Planar}} & \multicolumn{3}{c}{\cellcolor[HTML]{E2EFDA}\textbf{Volumetric}} \\
 & \textbf{Track} & \textbf{IoU} & \textbf{PSNR} & \textbf{Track} & \textbf{IoU} & \textbf{PSNR} & \textbf{Track} & \textbf{IoU} & \textbf{PSNR} \\ \midrule
\textbf{PhysTwin} \cite{Phystwin} & \textbf{0.014} & 54.51 & 27.29 & \textbf{0.036} & 68.01 & 24.16 & 0.026 & 70.35 & 24.15 \\
\textbf{PhysReal} & \textbf{0.014} & \textbf{70.17} & \textbf{30.46} & 0.041 & \textbf{75.70} & \textbf{25.30} & \textbf{0.021} & \textbf{77.56} & \textbf{25.46} \\
\bottomrule
\end{tabular}
\vspace{-5pt}
}
\\
\end{table}

\textbf{Analysis across Object Geometries.}
\textcolor{black}{Tab.~\ref{table:type} shows that PhysReal achieves higher IoU and PSNR across all geometries on the PhysTwin dataset, with Track Error matching PhysTwin on linear objects, improving on volumetric objects, but worsening on planar objects. Cloth resists stretching yet bends readily, while our volumetric MPM represents bending through strain variations across its thickness. Limited thickness resolution may therefore impair bending and local trajectory prediction. This limitation also helps explain why PhysReal's overall prediction Track Error matches PhysTwin's in Tab.~\ref{table:baseline_compare}.}

\subsection{Ablation Studies}
\label{sec:ablation}

\textcolor{black}{We select three representative sequences, Push rope, Lift cloth, and Lift zebra, covering linear, planar, and volumetric objects, respectively. Tab.~\ref{table:ablation_loss} evaluates the constitutive model and learning strategy on held-out prediction frames.}

\textcolor{black}{\textbf{Constitutive Model.}
The homogeneous variant shares $E$, $\nu$, and both elastic and plastic latent codes across all patches while retaining both neural residual branches. Its substantially worse tracking and visual fidelity highlight the importance of spatially varying constitutive modeling. Removing either or both residual branches also lowers mean PSNR, with the relative importance of elastic and plastic corrections varying across objects. Replacing constitutive residuals with additive velocity corrections slightly lowers mean Track Error but reduces PSNR. This trade-off suggests that state corrections favor fitting sparse trajectories, whereas constitutive corrections better preserve overall deformation and visual fidelity.}

\begin{figure}[!t]
    \centering
    \includegraphics[width=\columnwidth]{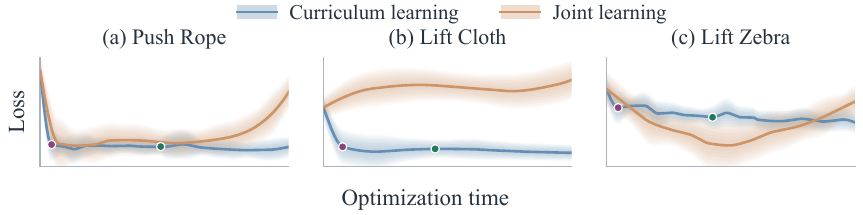}
    \caption{\textcolor{black}{Curriculum and joint optimization of the same complete model under equal wall-clock budgets. For each task, both methods' losses are normalized by the same maximum loss. Shading indicates the standard deviation across five random seeds. Colored dots mark curriculum stage transitions.}}
    \label{fig:optimization_comparison}
    \vspace{-5pt}
\end{figure}

\textcolor{black}{\textbf{Learning Strategy.}
Removing either $\mathcal{L}_{\mathrm{2D}}$ or $\mathcal{L}_{\mathrm{3D}}$ has a limited effect, consistent with their shared tracked correspondences. Removing all motion supervision particularly harms cloth, whose internal folds cannot be resolved from silhouettes alone. Removing mask supervision instead strongly affects rope, whose silhouette constrains its global shape. Joint optimization of the same complete model also degrades prediction performance. Fig.~\ref{fig:optimization_comparison} shows greater instability under joint optimization: loss can rise after initial progress, with larger variation across repeated runs. In contrast, the curriculum progressively optimizes coupled parameter groups, yielding smoother convergence and lower final losses.}

\begin{table}[!t]
\centering
\caption{\textcolor{black}{Component and optimization ablations on future prediction frames of three representative sequences.}}
\vspace{-5pt}
\label{table:ablation_loss}
\begingroup
\color{black}
\scriptsize
\renewcommand\arraystretch{1.0}
\setlength{\tabcolsep}{1.5pt}
\resizebox{\columnwidth}{!}{%
\begin{tabular}{clcccccccc}
\toprule
 & & \multicolumn{2}{c}{\cellcolor[HTML]{E2EFDA}\textbf{Push Rope}} & \multicolumn{2}{c}{\cellcolor[HTML]{D9E1F2}\textbf{Lift Cloth}} & \multicolumn{2}{c}{\cellcolor[HTML]{E2EFDA}\textbf{Lift Zebra}} & \multicolumn{2}{c}{\textbf{Mean}} \\
 & & Track & PSNR & Track & PSNR & Track & PSNR & Track & PSNR \\
\midrule
\multirow{5}{*}{\rotatebox[origin=c]{90}{\shortstack{Constitutive\\model}}}
 & w/o heterogeneity & 0.023 & 24.02 & 0.042 & 23.22 & 0.021 & 23.00 & 0.029 & 23.41 \\
 & w/o $\mathcal{N}_e$ & \underline{0.019} & 28.99 & 0.040 & 22.95 & \underline{0.019} & 24.98 & \underline{0.026} & 25.64 \\
 & w/o $\mathcal{N}_p$ & \underline{0.019} & 28.07 & 0.040 & 23.11 & 0.020 & 23.73 & \underline{0.026} & 24.97 \\
 & w/o $\mathcal{N}_e,\mathcal{N}_p$ & \underline{0.019} & 29.00 & 0.039 & 23.27 & 0.024 & 23.40 & 0.027 & 25.22 \\
 & State residuals & \underline{0.019} & 28.90 & \underline{0.037} & 22.59 & \underline{0.019} & 24.14 & \textbf{0.025} & 25.21 \\
\midrule
\multirow{5}{*}{\rotatebox[origin=c]{90}{\shortstack{Learning\\strategy}}}
 & w/o $\mathcal{L}_{\mathrm{2D}}$ & \textbf{0.017} & \underline{29.81} & 0.042 & 23.27 & \underline{0.019} & 25.12 & \underline{0.026} & \underline{26.07} \\
 & w/o $\mathcal{L}_{\mathrm{3D}}$ & 0.020 & 29.49 & 0.040 & 23.06 & \textbf{0.018} & 24.92 & \underline{0.026} & 25.82 \\
 & w/o $\mathcal{L}_{\mathrm{motion}}$ & \textbf{0.017} & 29.00 & 0.056 & 23.22 & \underline{0.019} & 24.96 & 0.031 & 25.73 \\
 & w/o $\mathcal{L}_{\mathrm{mask}}$ & 0.036 & 24.60 & \textbf{0.035} & \textbf{23.37} & 0.023 & 24.68 & 0.031 & 24.22 \\
 & Joint optimization & 0.020 & 27.73 & 0.070 & 20.18 & 0.023 & \underline{25.34} & 0.038 & 24.42 \\
\midrule
\multicolumn{2}{l}{\textbf{PhysReal}} & \underline{0.019} & \textbf{29.90} & 0.039 & \underline{23.28} & \underline{0.019} & \textbf{25.39} & \underline{0.026} & \textbf{26.19} \\
\bottomrule
\end{tabular}%
\vspace{-5pt}
}
\endgroup
\end{table}

\begin{figure}[!t]
    \centering
    \includegraphics[width=0.95\columnwidth]{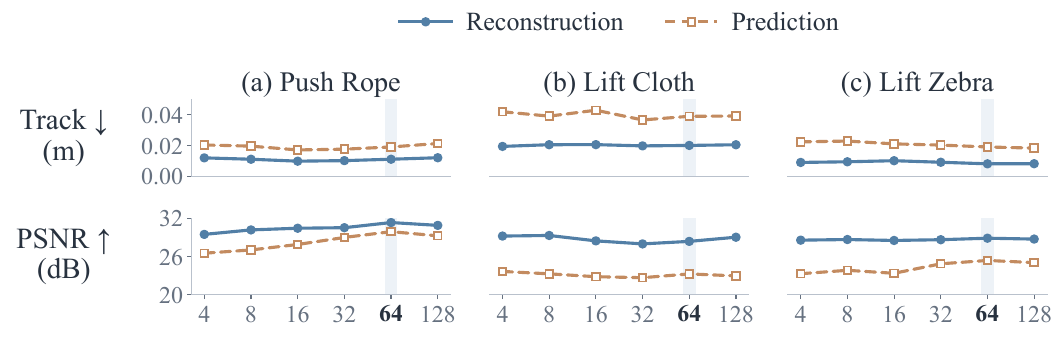}
    \caption{\textcolor{black}{Effect of patch count on reconstruction and prediction. Solid lines indicate reconstruction on training frames, while dashed lines indicate prediction on test frames.}}
    \label{fig:patch_count_effect}
    \vspace{-5pt}
\end{figure}

\textcolor{black}{\textbf{Effect of Patch Count.}
Fig.~\ref{fig:patch_count_effect} shows how the number of patches affects reconstruction and prediction performance. Despite their different geometries and corresponding patch resolutions, reconstruction and prediction remain broadly stable. Increasing the patch count provides modest, object-dependent benefits. Finer spatial resolution can help distinguish local material responses, as needed in Push rope, where the contacted region bends while other regions resist excessive stretching. We use 64 patches for all objects as a practical accuracy-efficiency compromise, while allowing the count to be adjusted to the desired spatial resolution and computational budget.}

\subsection{\textcolor{black}{Recovered Material Properties}}
\label{sec:material_properties}

\begin{figure}[!t]
    \centering
    \includegraphics[width=0.9\columnwidth]{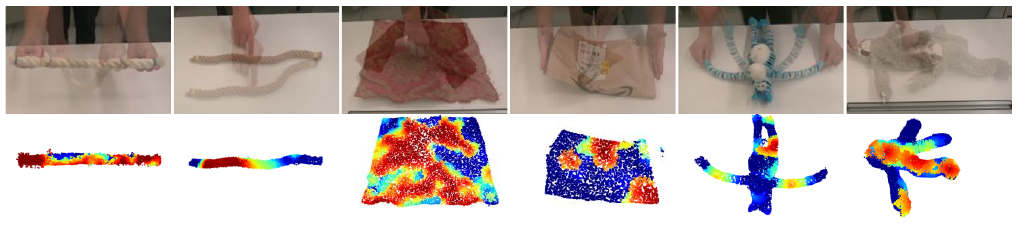}
    \caption{\textcolor{black}{3D representations colored by the recovered Young's modulus. Colors range from red for higher values to blue for lower values.}}
    \label{fig:material_properties}
    \vspace{-5pt}
\end{figure}

\textcolor{black}{To examine whether the recovered material parameters reflect physically meaningful local compliance, we visualize Young's modulus $E$ on the 3D particles of representative objects in Fig.~\ref{fig:material_properties}. Regions that undergo pronounced bending during the observed interactions tend to have lower recovered moduli, while regions that remain relatively rigid tend to have higher values. These patterns suggest that the identified material properties provide a physically plausible explanation of the object deformation under the observed interaction.}

\begin{figure}[!t]\centering
	\includegraphics[width=0.95\columnwidth]{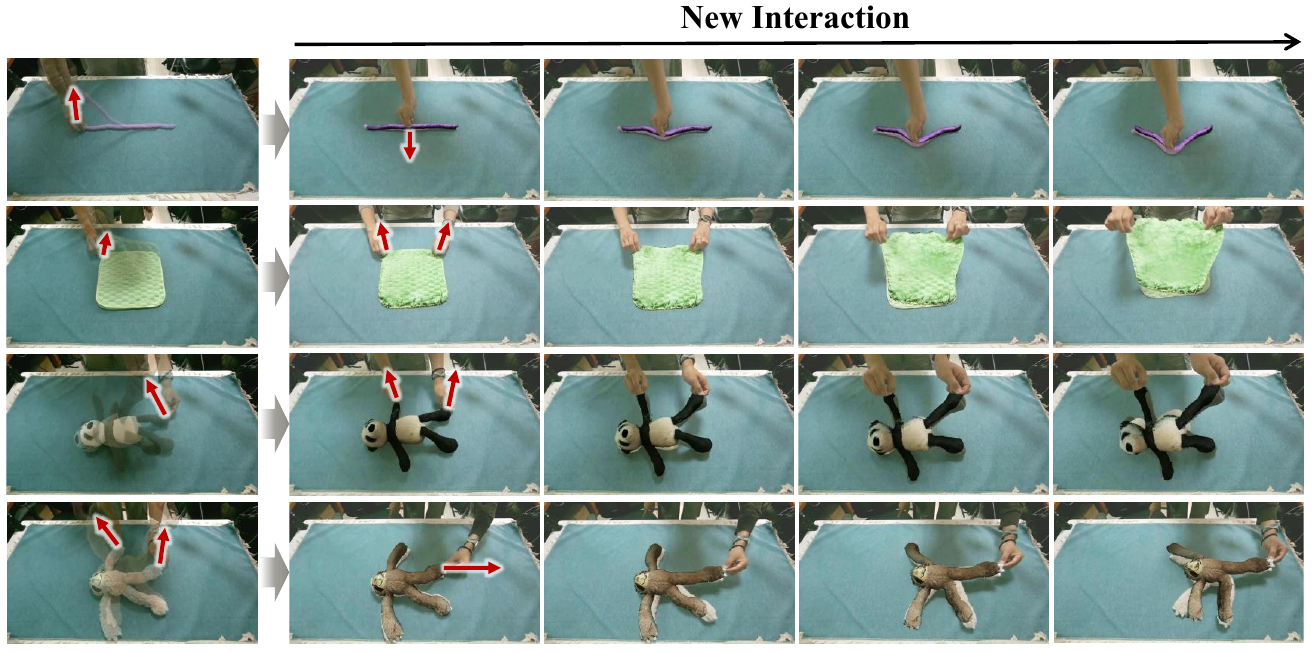}
    \caption{\textcolor{black}{Cross-interaction generalization on our self-collected objects. Left: source interactions used for identification. Right: rollouts under different target interactions without further optimization.}}
    \label{fig:generalization}
    \vspace{-5pt}
\end{figure}

\begin{figure}[!t]
    \centering
    \includegraphics[width=\columnwidth]{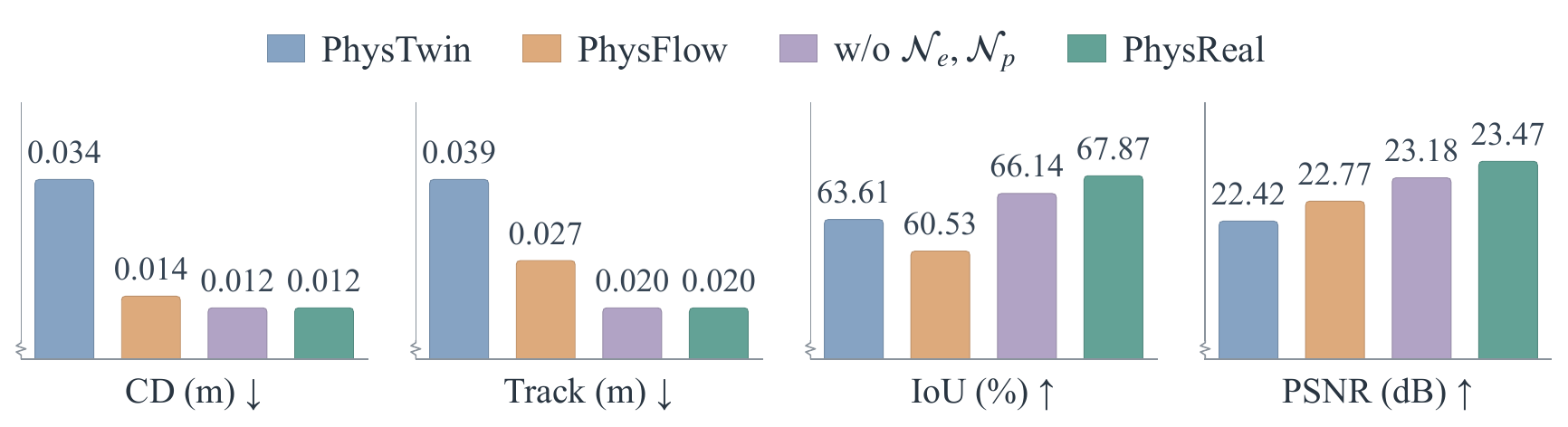}
    \caption{\textcolor{black}{Quantitative cross-interaction results averaged over four objects, with one source-to-target transfer per object.}}
    \label{fig:cross_interaction_metrics}
    \vspace{-5pt}
\end{figure}

\subsection{Generalization to Unseen Interactions}
\label{sec:generalization}

\textcolor{black}{We choose our self-collected dataset for this evaluation because different interaction sequences of the same object start from aligned configurations. This allows the identified geometry and material field to be reused directly, without additional deformation to match a new initial shape. We evaluate one source-to-target transfer for each of the four objects. Each model is identified from the source sequence and then rolled out under the target interaction trajectories with all learned parameters fixed and no target-sequence fitting. Fig.~\ref{fig:generalization} illustrates the transferred dynamics of PhysReal.}

\textcolor{black}{Fig.~\ref{fig:cross_interaction_metrics} compares PhysReal with PhysTwin, PhysFlow, and a spatially varying expert-only variant without either neural residual branch. PhysReal achieves lower geometric and tracking errors and higher visual fidelity than PhysTwin and PhysFlow. PhysTwin's larger tracking errors may reflect limited transfer of its fitted spring response to new loads. PhysReal instead uses a continuum constitutive prior to model responses under different loading conditions. Neural residuals modestly improve visual fidelity over the expert-only variant, with comparable CD and Track Error. The gains are smaller than in Sec.~\ref{sec:ablation}, possibly because a single source interaction covers too few target deformation states for the learned corrections to transfer fully.}

\subsection{Downstream Applications}
\label{sec:applications}

\textcolor{black}{We explore potential downstream uses of the recovered physical model through qualitative demonstrations, as illustrated in Fig.~\ref{first}. The model supports interactive simulation under user-specified trajectories. Its particle states provide grasping and interaction targets, while 3DGS rendering supplies paired visual observations for data generation. Integrating the recovered object with the reconstructed environment and robot further provides a physical digital twin for visualizing workspace interactions. The identified dynamics also serve as a forward model for model predictive control (MPC), which evaluates candidate actions to plan toward a target object configuration. Preliminary timing measurements indicate approximately 10\,ms per MPM substep and 10\,ms per rendered frame, while MPC-based trajectory optimization takes several minutes per planning run.}

\section{\textcolor{black}{Limitations}}
\label{sec:limitations}

\textcolor{black}{PhysReal currently learns from a single interaction video, limiting the range of deformation states observed during identification. Its cross-interaction rollouts can therefore match observations less closely than models fitted to the target interaction. Thin planar objects such as cloth also remain challenging, with ambiguous visual supervision of folds and limited resolution of membrane and bending responses in our volumetric MPM formulation.}

\section{Conclusion}
\label{sec:conclusion}
We presented PhysReal, a video-driven framework for learning deformable object dynamics through spatially varying constitutive modeling. By combining expert physics priors with neural residuals in a differentiable MPM simulator, PhysReal recovers complex material behaviors from sparse interactive observations.
Experiments demonstrate that PhysReal achieves more accurate reconstruction and future prediction than existing approaches. \textcolor{black}{Preliminary demonstrations also explore potential robotic applications.}
Future research will explore jointly learning from multiple interaction videos, where diverse manipulation trajectories can provide complementary observations of the underlying physical properties.

\bibliographystyle{Bibliography/IEEEtran}
\bibliography{physreal}

\end{document}